# Deep Learning and Word Embeddings for Tweet Classification for Crisis Response

Reem ALRashdi
Department of Computer Science; Department of Computer Science and Software Engineering
University of York; University of Hail
York, United Kingdom; Hail, Saudi Arabia
Rmma502@york.ac.uk; r.alreshede@uoh.edu.as

Simon O'Keefe
Department of Computer Science
University of York
York, United Kingdom
Simon.okeefe@york.ac.uk

*Abstract*—Tradition tweet classification models for crisis response focus on convolutional layers and domain-specific word embeddings. In this paper, we study the application of different neural networks with general-purpose and domain-specific word embeddings to investigate their ability to improve the performance of tweet classification models. We evaluate four tweet classification models on CrisisNLP dataset and obtain comparable results which indicates that general-purpose word embedding such as GloVe can be used instead of domain-specific word embedding especially with Bi-LSTM where results reported the highest performance of 62.04% F1 score.

## I. INTRODUCTION

Twitter has become a dominant platform for organizations and people to post or gather information during crises [1]. People spread the news on Twitter and share valuable, real-time and on-topic information like their statuses, injured or dead people and the damage caused by the crisis [2]. They also tweet to ask for help for themselves or offer help to others.

Twitter also has proven to be a powerful information source in many natural or human-made crisis situations during recent years such as earthquakes [3], floods [4], wildfires [5] and nuclear disasters [6]. For example, in 2011, 177 million crisis-related tweets were published in only one day during an earthquake in Japan [7]. Another example is when a haze hit Singapore in 2013 where people posted more than 23 million informative tweets [8].

It is obvious that situational awareness can be significantly enhanced by people-generated tweets [2]. These tweets can be used by large-scale disaster response organizations to make better decisions and quick response. However, humanitarian organizations aim at responding to people in help cannot manually observe, process and convert the enormous volume of information into actionable one [9]. Thus, they do not widely use social media data such as Twitter in their disaster response operations [10].

Deep neural networks have proven their ability to automatically learn deep and complicated mappings from input to output by using distributed representation of words without requiring any feature engineering. It is also noticeable that deep learning approaches have outperformed traditional ones in many Natural Language Processing (NLP) tasks including tweet classification [11]. Tweet classification for crisis response is a text classification task that aims at identifying if a tweet is related to a specific type of predefined informative classes.

In addition, a powerful word embedding can be a key factor in improving the neural network performance in any NLP task [1]. General-purpose and domain-specific word embeddings have been proposed recently such as Global Vectors (GloVe) embedding [12] and Crisis embedding [11]. However, few numbers of experiments have been conducted to examine the effectiveness of different deep learning architectures and different word embeddings in improving tweet classification models.

Our work is similar to [1] and [11] however we use different neural network architectures and different general-purpose word embeddings. We also train our four models separately in an offline fashion without integrating any network components. To the best of our knowledge, this is the first study of using GloVe embedding with Bidirectional Long Short-Term Memory (Bi-LSTM) or Convolutional Neural Network (CNN) in the field of tweet classification for crisis response.

The input of our networks are tweets that may contain any information related to any natural crisis. In this paper, we target natural crises such as earthquakes, floods, storms, typhoons and so on. These crisis types have specific properties: (1) people's daily life is affected when at least one of them occur, (2) they considered to be large-scale events because big number of people experience them, and (3) every crisis has an associated time and location. First of all, we clean the tweets by removing unnecessary parts such as emojis and http addresses. Then, the tweets are tokenized into words and a pre-trained word embedding (GloVe or Crisis) is used to capture similarities between words and semantics of word sentences. After that, a deep learning architecture (CNN or Bi-LSTM) is applied to encode and leverage the information from the input text sequence, tweets. Finally, a fully connected layer with a softmax layer are used to compute the class distribution for each tweet.

## II. Deep Learning Architectures

### A. Convolutional Neural Networks

CNN is a deep learning architecture that consists of an input layer, multiple neural hidden layers and an output layer. Usually in NLP tasks, token sequences are used as input to the CNN. Then, CNN filters preform as n-grams over continuous representations. After that, these n-grams filters are combined by subsequent network layers, dense layers [12].

CNN can learn the features and distinguish between them automatically and therefore CNN does not require hand-engineered features which saves human effort and time and eliminates the need of prior knowledge. And unlike a multilayer perceptron (MLP), the number of free parameters can be reduced by CNNs and the vanished or exploded gradients can be prevented during the training process. Also, all the weights in the convolutional layers are shared which means that the same filter is used for all the fields within a layer to improve the performance and decrease the memory space

### B. Bidirectional Long Short-Term Memory

Long Short-Term Memory networks (LSTMs) are variants of Recurrent Neural Networks (RNNs) that solve the gradient vanishing/exploding problems of RNNs [13]. Basically, LSTMs are designed to capture long-distance dependencies within texts. Each LSTM unit consists of three gates to control which portions of information to remember, forget and pass to the next step. LSTMs hold the contextual semantics of each word by the surrounding information and store long dependencies between words. However, they only focus on one direction of the input which is the past.

On the other hand, Bi-LSTMs focus on the past and the future directions of the input. This method allows the network to capture more information than before where at every token position, hidden representations from each direction are concatenated.

## III. Word Embeddings

The application of word embedding has been drawn great interest in NLP during the last few years. Word embedding is a set of feature learning techniques or language models where texts (phrases or words) are mapped to real numbers vectors. The main goal of word embedding is to learn efficient and expressive text representations where similar words or phrases have similar representations that capture their semantic meaning [14].

## IV. Network Training

In this section, we provide details about training four classifiers: CNN with Crisis embedding, CNN with GloVe embedding, Bi-LSTM with Crisis embedding and Bi-LSTM with GloVe embedding.

### A. Tweets Pre-processing

Tweets are full of noise because of the presence of incomplete sentences or words, irregular expressions, ill-formed sentences or words and out of dictionary words. Thus, we clean all input tweets by removing all the http addresses, hashtags, emojis, stop words and punctuations.

### B. Word Embedding Initialization

We use two kinds word embeddings to initialize the embeddings in the start of all the experiments: GloVe embedding and Crisis embedding.

GloVe embedding is a very known general pretrained word embedding created by the authors in [15]. It has been proven that this embedding played a key role in improving many NLP tasks [15]. GloVe embedding is a publicly available 100-dimentional embedding trained on 6 billion words from web text and Wikipedia which is similar to social media texts such as tweets.

Unlike GloVe embedding, Crisis embedding is a domain-specific pretrained word embedding founded by [11]. The 300-dimentional embedding trained on 20 million words from 57,908 disaster related tweets which is the desirable domain.

### C. Hyper-parameters Intialization

Table 1 demonstrates the values of the selected hyper-parameters for all the four experiments.

TABLE I. Hyper-parameters for all experiments.

| Layer | Hyper-parameters | Values |
|---|---|---|
| CNN | Kernel size | 3 |
| CNN | Pool size | 2 |
| CNN | Number of filters | 250 |
| CNN | Hidden size | 128 |
| Bi-LSTM | Hidden size | 100 |
|  | Batch size | 32 |
|  | Epoch | 25 |

### D. Fine Tuning

The initial embedding of the GloVe embedding has been fine-tuned by updating them during the gradient modification of the deep learning model using back-propagating gradients. Fine-tuning word embedding represents transferring the knowledge from the initial corpus where the embedding was built to our domain dataset. On the other hand, we have not fine-tuned the initial embedding of the Crisis embedding since it was trained on crisis-related tweet corpus and no significant improvement was reported in such experiment in previous papers.

In addition, it is infeasible to fine tune all hyper-parameters by random search in all the experiments due to time constraints. However, we have followed the authors in [11] in choosing the hyper-parameters values at the start of all the experiments.

### E. Dropout Training

We apply dropout on word embedding before inputting to the deep learning architectures and on the input and the output of each model where each node is removed with the probability of *1 - p* or kept with the probability of *p* only in the training time to avoid training all the nodes. As expected, the models' performance significantly improved after using dropout which proves the effectiveness of dropout in reducing models' overfitting.

## V. EXPERIMENTS

### A. Models

We conduct four experiments using different word embeddings (crisis embedding and Glove) and deep learning architectures (CNN and Bi-LSTM). We have re-implemented the CNN and Crisis embedding model from [11] to compare it with the other three models in order to investigate the effectiveness of integrating different word embeddings with different deep learning architectures. Fig. 1 describes the first and the second classifiers where CNN is used with GloVe and Crisis embedding separately.

Fig. 1. Conventional Nueral Network *(CNN)* with word embedding for twitter classification for crisis response

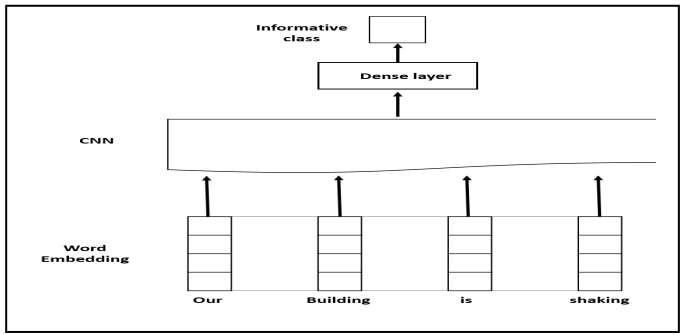

We use Bi-LSTM in the third experiment with GloVe embedding and with Crisis embedding in the fourth experiment as shown in Fig. 2.

Fig. 2. Bidirectional Long Short-Term Memory *(Bi-LSTM)* with word embedding for twitter classification for crisis response.

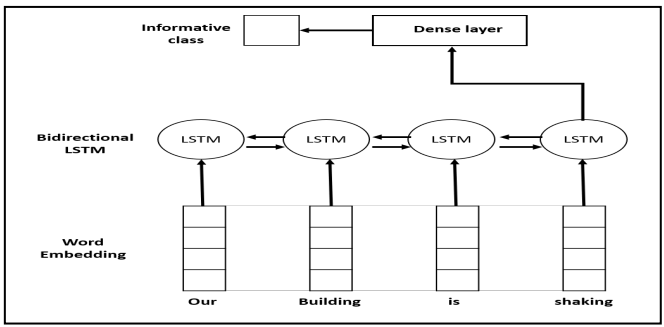

### B. Data sets

We use CrisisNLP dataset [16] to evaluate the four classifiers mentioned in the previous section. CrisisNLP is a collection of small datasets where each dataset contains annotated tweets related to a crisis event. The tweets are labelled based on their corresponding informative class (e.g. affected individuals, donations and volunteering, infrastructure and utilities, sympathy and support, other useful information and irrelevant). Number of tweets for each set is shown in Table 2.

TABLE II. DISTRIBUTION OF TWEETS DURING NETWORKS TRAINING AND TESTING

| Class number | Class title | Number of tweets | | |
|---|---|---|---|---|
| | | Train set | Dev set | Test set |
| 1 | Injured of dead people | 1611 | 487 | 233 |
| 2 | Missing, trapped or found people | 741 | 221 | 106 |
| 3 | Infrastructure and utilities damages | 676 | 177 | 94 |
| 4 | Sympathy and emotional support | 1526 | 436 | 232 |
| 5 | Donation needs or offers or volunteering services | 2352 | 712 | 350 |
| 6 | Other useful information | 5690 | 1623 | 766 |
| 7 | Irrelevant | 6254 | 1756 | 886 |
| | Total number of tweets | 18850 | 5412 | 2667 |

### C. Evaluation Metrics

We use F1 score to evaluate and compare the models due to the imbalanced dataset. F1 score is calculated with the formula in (1) and the final value relies between 0 and 1 where 1 indicates a perfect model.

$$F1\ score = 2 * \frac{1}{\frac{TP+FP}{TP} + \frac{TP+FN}{TP}} \quad (1)$$

Where True Positives (TP) is the correctly predicted positive values, False Positive (FP) is the wrongly predicted positive values and False Negative (FN) is the wrongly predicted negative values.

### D. Results

CNN with Crisis embedding model achieved an F1 score of 61.38 which is slightly higher than the model that contains Bi-LSTM with the same embedding. Bi-LSTM with GloVe embedding reported the best result among all the four models with 62.04 F1 score and CNN with GloVe embedding recorded the worst performance with 59.87 F1 score. The results of all the experiments are shown below in Table 3.

TABLE III. RESULTS OF FOUR EXPERIEMENTS USING DIFFERENT DEEP LEAANING ARCHITECTURES ANS WORD EMBEDDINGS

| Experiment | Model components | | |
|---|---|---|---|
| | Deep Learning architecture | Word Embedding | F1-score |
| 1 | CNN | Crisis embedding | 61.38 |
| 2 | | GloVe embedding | 59.87 |
| 3 | Bi-LSTM | Crisis embedding | 60.88 |
| 4 | | GloVe embedding | 62.04 |

## VI. DISCUSSION

According to the results shown in the previous section, Bi-LSTM with GloVe model obtains the best performance for text classification for crisis response. That demonstrates the effectiveness of general pretrained word embedding such as GloVe and sequence models such as Bi-LSTM in improving the classifier ability to distinguish between crisis-related tweets.

However, domain-specific embedding outperforms general word embedding when integrated with CNN. This shows the importance of choosing a word embedding depending on the selected deep learning architecture for tweet classification.

We believe that the main reason behind these results is that Crisis embedding initially is built using the Skip-gram model of the Word2Vec tool, which is a very powerful method in detecting the semantic meaning of words with a small semantic space. On the other hand, the GloVe embedding needs more information than the Crisis embedding to successfully detect the semantic meaning of words. This is consistent with the fact that Bi-LSTM captures more information than CNN. Bi-LSTM captures the sequence of tweets in both directions while CNN captures the local patterns of tweets and may loses some information such as the words order in tweets.

Another possible reason is that Crisis embedding may contain twitter-specific text irregularities such as emojis, mentions hashtags and other domain-specific words. These were not taken under consideration when training GloVe embedding. Because we perform pre-processing for our dataset and remove such words the performances of both GloVe and Crisis embeddings are expected to be close. However, misspelling words such as 'flods' for 'floods' can only exist in Crisis embedding which gives it a very limited advantage over Glove embedding.

Finally, we discovered that GloVe as a general word embedding can be used instead of Crisis embedding as a domain-specific word embedding to improve the performance of Bi-LSTM-based model to classify tweets for crisis response.

## VII. RELATED WORK

Recently, a limited number of experiments has been reported on successfully applying deep learning architectures and word embeddings to tweet classification for crisis response. It started when the authors in [11] argued that the informative class in the previous studies still has a lot of information to be handled by organizations. To simplify the organizations' work and save their time and effort, they introduced a model that classified the informative class into multiple subclasses (e.g., infrastructure damages, affected people, donation and volunteering, sympathy and support and other useful information). This work is very similar to our first model where the authors build their model with a single CNN layer after a look-up layer and before a pooling layer. After that, a dropout layer is added to reduce the model's overfitting. However, they trained the initial model first and then retrained it with small mini-batches in an online fashion to suit the early crisis response situation where we use their pretrained word embedding (Crisis embedding) without retraining the model.

The same model (CNN and Crisis embedding) has been also used by the authors in [1] but they integrated the domain-specific word embedding with Google word embedding and results reported a slight improvement on the model's performance.

Another deep learning model has been introduced in [17]. The semantically-enhanced dual-CNN consists of two layers: a semantic layer that captures the contextual information and a traditional CNN layer. The results show that the dual-CNN model has a comparable performance with a single CNN.

## VIII. CONCLUSION AND FUTURE WORK

In this paper, we investigate the effect of using domain-specific and general word embeddings with two deep learning architectures: Bi-LSTM and CNN. Results reported that using different word embeddings slightly improve the model performance due to the variability of the corpora used when building the word embeddings.

Further experiments will be done to examine the effectiveness of N-Gram CNN, another architecture introduced in [12], in classifying tweets for crisis response. In addition, we will consider recent works in integrating general word embedding such as GloVe for rich semantic representations of general words and domain-specific embedding for domain-specific words such as ill-words within tweets in our case.

REFERENCES

[1] D. T. Nguyen,, K. A. A. Mannai, S. Joty, H. Sajjad, M. Imran, and P. Mitra, "Rapid Classification of Crisis-Related Data on Social Networks using Convolutional Neural Networks," 2016. [online]. Available: arXiv preprint arXiv:1608.03902. [Accessed Sep. 8, 2018]

[2] S. E. Vieweg, "Situational awareness in mass emergency: A behavioural and linguistic analysis of microblogged communications", Ph.D. dissertation, Univ. of Colorado, Boulder, 2012.

[3] Y. Qu, C. Huang, P. Zhang, and J. Zhang, "Microblogging after a major disaster in China: a case study of the 2010 Yushu earthquake". *In Proceedings of the ACM 2011 conference on Computer supported cooperative work* , *ACM, March, 2011*. pp. 25-34.

[4] K. Starbird, L. Palen, A. L. Hughes, and S. Vieweg, "Chatter on the red: what hazards threat reveals about the social life of microblogged information," In *Proceedings of the ACM 2010 conference on Computer supported cooperative work, ACM, February, 2010*. pp. 241-250.

[5] S. Vieweg, A. L. Hughes, K. Starbird and L. Palen, (2010, April). "Microblogging during two natural hazards events: what twitter may contribute to situational awareness," In *Proceedings of the SIGCHI conference on human factors in computing systems.ACM. April, 2010*. pp. 1079-1088.

[6] R. Thomson, N. Ito, H. Suda, F. Lin, Y. Liu, R. Hayasaka, R. Isochi and Z. Wang, " Trusting tweets: The Fukushima disaster and information source credibility on twitter," *In Proceedings of the 9th International ISCRAM Conference, ISCRAM, 2012*. pp. 1–10.

[7] S. E. Cho, K. Jung and H. W. Park, " Social media use during Japan's 2011 earthquake: how Twitter transforms the locus of crisis communication, " *Media International Australia,* vol. 149, ED-1, pp. 28-40. 2013.

[8] P. K. Prasetyo, M. Gao, E. P. Lim, and C. N. Scollon, " Social sensing for urban crisis management: The case of singapore haze," *In International Conference on Social Informatics, Springer, Cham, November, 2013*. pp. 478-491.

[9] H. Gao, G. Barbier and R. Goolsby, "Harnessing the crowdsourcing power of social media for disaster relief,". *IEEE Intelligent Systems*, vol. 26, ED-3, pp. 10-14. 2011.

[10] A. H. Tapia and K. Moore, "Good enough is good enough: Overcoming disaster response organizations' slow social media data adoption", *Computer Supported Cooperative Work (CSCW)*, vol. 23, ED-(4-6), pp. 483-512. 2014.

[11] D. T. Nguyen, S. Joty, M. Imran, H. Sajjad and P. Mitra, "Applications of online deep learning for crisis response using social media information," 2016. [online]. Available: arXiv preprint arXiv:1610.01030. [Accessed Sep. 8, 2018].

[12] Y. Kim, "Convolutional neural networks for sentence classification," 2014. [online]. Available: arXiv preprint arXiv:1408.5882. [Accessed Sep. 8, 2018].

[13] S. Hochreiter and J. Schmidhuber, "Long short-term memory," *Neural computation,* vol. 9, ED-8, pp.1735-1780. 1997.

[14] M. Naili, A. H. Chaibi and H. H. B. Ghezala, "Comparative study of word embedding methods in topic segmentation," *Procedia Computer Science*, vol. 112, pp. 340-349.

[15] J. Pennington, R. Socher and C. Manning, "Glove: Global vectors for word representation," *In Proceedings of the 2014 conference on empirical methods in natural language processing (EMNLP), 2014.* pp. 1532-1543.

[16] M. Imran, P. Mitra and C. Castillo, "Twitter as a lifeline: Human-annotated twitter corpora for NLP of crisis-related messages,". 2016. [online]. Available: arXiv preprint arXiv:1605.05894. [Accessed Sep. 8, 2018].

[17] G. Burel and H. Alani, "Crisis Event Extraction Service (CREES)- Automatic Detection and Classification of Crisis-related Content on Social Media," *In the 15th International Conference on Information Systems for Crisis Response and Management, 20-23 May 2018, Rochester, NY, USA.*